\documentclass[nohyperref]{article}

\usepackage{microtype}
\usepackage{graphicx}
\usepackage{subfigure}
\usepackage{booktabs} 

\usepackage{hyperref}

\usepackage[accepted]{icml2022}

\usepackage{microtype}
\usepackage[utf8]{inputenc} 
\usepackage[T1]{fontenc}    
\usepackage{hyperref}       
\usepackage{url}            
\usepackage{booktabs}       
\usepackage{amsfonts}       
\usepackage{nicefrac}       
\usepackage{microtype}      
\usepackage{xcolor}         
\usepackage{multirow}
\usepackage{enumitem}
\usepackage{bm}
 \usepackage{graphicx}
\usepackage{graphics}
\usepackage{wrapfig}
\usepackage{amsmath}
\usepackage{amsthm}
\usepackage{caption}
\usepackage{color}
\usepackage{bbm}

\newcommand{\eat}[1]{}
\newcommand{\calG}{\mathcal{G}}

\icmltitlerunning{Analyzing and Mitigating Interference in Neural Architecture Search}

\begin{document}

\twocolumn[
\icmltitle{Analyzing and Mitigating Interference in Neural Architecture Search}




\begin{icmlauthorlist}
\icmlauthor{Jin Xu}{yyy}
\icmlauthor{Xu Tan}{comp}
\icmlauthor{Kaitao Song}{comp}
\icmlauthor{Renqian Luo}{comp}
\icmlauthor{Yichong Leng}{comp,sch}
\icmlauthor{Tao Qin}{comp}
\icmlauthor{Tie-Yan Liu}{comp}
\icmlauthor{Jian Li}{yyy}
\end{icmlauthorlist}

\icmlaffiliation{yyy}{Institute for Interdisciplinary Information Sciences (IIIS), Tsinghua University}
\icmlaffiliation{sch}{University of Science and Technology of China}
\icmlaffiliation{comp}{Microsoft Research Asia}

\icmlcorrespondingauthor{Xu Tan}{xuta@microsoft.com}
\icmlcorrespondingauthor{Jian Li}{lijian83@mail.tsinghua.edu.cn}

\icmlkeywords{Machine Learning, ICML}

\vskip 0.3in
]



\printAffiliationsAndNotice{}  

\begin{abstract}
Weight sharing is a popular approach to reduce the cost of neural architecture search (NAS) by reusing the weights of shared operators from previously trained child models. However, the rank correlation between the estimated accuracy and ground truth accuracy of those child models is low due to the interference among different child models caused by weight sharing. In this paper, we investigate the interference issue by sampling different child models and calculating the gradient similarity of shared operators, and observe: 1) the interference on a shared operator between two child models is positively correlated with the number of different operators; 2) the interference is smaller when the inputs and outputs of the shared operator are more similar. Inspired by these two observations, we propose two approaches to mitigate the interference: 1) MAGIC-T: rather than randomly sampling child models for optimization, we propose a gradual modification scheme by modifying one operator between adjacent optimization steps to minimize the interference on the shared operators; 2) MAGIC-A: forcing the inputs and outputs of the operator across all child models to be similar to reduce the interference. Experiments on a BERT search space verify that mitigating interference via each of our proposed methods improves the rank correlation of super-net and combining both methods can achieve better results. Our discovered architecture outperforms RoBERTa$_{\rm base}$ by 1.1 and 0.6 points and ELECTRA$_{\rm base}$ by 1.6 and 1.1 points on the dev and test set of GLUE benchmark. Extensive results on the BERT compression, reading comprehension and ImageNet task demonstrate the effectiveness and generality of our proposed methods.
\end{abstract}

\section{Introduction}
Neural Architecture Search (NAS) \citep{zoph2017neural,zoph2018learning,liu2018darts} aims to automatize the process of discovering high-performance architectures.
Conventional NAS~\cite{zoph2017neural,zoph2018learning,real2017large} obtains the accuracy of an architecture by training it from scratch, which requires huge computational resources (e.g., hundreds of GPU days). To speed up the training process, an important technique, called weight sharing, is proposed by~\citet{pham2018efficient}. Briefly speaking, weight sharing is to maintain a single copy of weights on a super-net that contains all the architectures in the search space. Rather than stand-alone training from scratch for each architecture, the weight sharing method samples an architecture at each training step as a child model or a sub-graph of the super-net, and continues the training on the weights of shared operators from previously trained architectures in the super-net. Due to low computational cost and competitive performance, weight sharing has drawn wide attention~\cite{pham2018efficient,bender2018understanding,liu2018darts,XieZLL19,cai2018proxylessnas,cai2019once, li2020random,guo2020single,peng2020cream,chu2019fairnas, ning2021evaluating}.

Although a number of NAS methods based on weight sharing have been proposed and achieved impressive results,  many works~\cite{ning2021evaluating,guo2020single,zela2019bench,yu2019evaluating,zhang2020deeper,zhang2020does} find that the rank correlation between the estimated accuracy and ground truth accuracy of child models is low due to the interference among different child models on shared weights. 
The shared operators receive different gradient directions from child models with different architecture topologies for optimization even with the same batch of training data, which seriously affects the rank of the child models since interference can cause insufficient training and inaccurate performance evaluation. This phenomenon is especially severe for the complex search space that contains different types of candidate operators (e.g., convolutions and multi-head attention) and numerous child models.

While previous research works have noticed the interference issue and empirically found that the interference is correlated with factors such as the size of a search space~\cite{shu2019understanding,zhang2020deeper}, little has been discussed about the causes of the interference and how to mitigate it. Through our quantitative analyses, we observe the interference on a shared operator is positively correlated with the number of different operators that exist between two child models. The main reason for the interference is that the topology of the architecture varies randomly along with the training process. Then the operator shared by different architecture topologies may receive activations from different inputs during the forward process and receive gradients from different outputs during the backward process. In this way, the shared operator is optimized towards different directions by different child models, which causes interference. Thus, we force the inputs and outputs of the operator to be similar to the average inputs and outputs of all the child models, and find that the interference can be reduced.

Inspired by the above observations, we propose two methods for \underline{M}itig\underline{A}tin\underline{G} \underline{I}nterferen\underline{C}e (MAGIC) from different perspectives: 1) reduce the changes of the architecture topologies in the adjacent sampling steps (MAGIC-T) and 2) align the inputs and outputs of the operator shared by different child models (MAGIC-A). As for MAGIC-T, we first analyze the commonly-used random single path one-shot algorithms~\cite{guo2020single} and find that the number of different operators between adjacent sampling steps is positively correlated with the number of layers of the super-net and number of candidate operators, which leads to serious interference in a large search space. 
To minimize the interference, we gradually change the topological environment of the shared operators by sampling a child model that differs from the child model sampled at the previous step with only one operator at each training step. As for MAGIC-A, we select the model with the best validation accuracy among the child models as an anchor model to align the inputs and outputs of all child models together respectively. The anchor model can be replaced when the performance of any child model outperforms it. Finally, MAGIC-T and MAGIC-A can be combined to further reduce the interference.

To verify the effectiveness of our methods, we adopt a challenging hybrid super-net including three types of operators: multi-head attention, feed-forward network and convolution, and conduct experiments on the large scale BERT pre-training task. The experiments 
verify that MAGIC-T and MAGIC-A 
can mitigate the interference and improve rank correlation individually. Combining them can achieve better performance. Extensive experiments on 10 natural language processing datasets 
of BERT pre-training, and ImagetNet classification tasks on MobileNet-V2 search space~\cite{sandler2018mobilenetv2} show the effectiveness and generality of our proposed methods. 

The contributions of this paper are summarized as follows:
\begin{itemize}[leftmargin=*]
\item We conduct thorough analyses on the interference issue in NAS, and find interference between two models is caused by diverse gradient directions, and is positively correlated with differences of their architecture topologies as well as their inputs and outputs of shared operators. To the best of our knowledge, we are the first to quantitatively analyze the 
interference with thorough empirical results. 
\item Inspired by our analyses, we propose two methods, MAGIC-T and MAGIC-A, to mitigate interference by reducing topological changes during the sampling process and aligning the inputs and outputs of the shared operators among different child models.
\item Experiments on the BERT search space demonstrate that our methods can significantly improve the rank correlation of the super-net by mitigating interference. Our discovered architecture outperforms RoBERTa$_{\rm base}$ by 1.1 and 0.6 points, and ELECTRA$_{\rm base}$ by 1.6 and 1.1 points on the dev and test of GLUE tasks respectively. Extensive experiments on the BERT compression task, reading comprehension and ImageNet classification tasks verify the generality of our proposed methods.
\end{itemize}

\section{Related Work}
\paragraph{Neural architecture search.}
Recently, NAS methods~\cite{pham2018efficient,xu2019pc, ning2021evaluating} mainly adopt weight sharing to re-use the weights of previously trained architectures within a super-net (a.k.a one-shot NAS methods) to speed up the training process. Many methods leverage differentiable NAS~\cite{liu2018darts,dong2019searching,wang2021rethinking} by relaxing the discrete architectures into continuous space, and jointly learn the super-net weights and architectural weights. 
A number of other approaches utilize sampled single path one-shot NAS
approaches~\cite{bender2018understanding,chu2019fairnas,li2020random,guo2020single,cai2019once}. They usually adopt a chain-styled super-net, train it by randomly sampling and optimizing a single path (child model), and then use the trained super-net as a performance estimator to evaluate the accuracy of child models to search for good architectures using progressively shrinking~\cite{xu2021taskagnostic,HuLGWZWGS20Angle} or evolution algorithms~\cite{guo2020single,yu2020bignas}. They enjoy the flexibility of decoupling the training and searching stages and can support searching multiple efficient networks under different constraints~\cite{cai2019once,xu2021taskagnostic}. However, in the sampling process, different optimization directions, caused by training different child models, interfere with each other, which causes insufficient training and inaccurate performance evaluation~\cite{zhang2020deeper}. Our work focuses on the interference issue of chain-styled search space in sampled single path one-shot NAS approaches.

\paragraph{Weight sharing and interference.}
Multiple works have analyzed the effects of weight sharing. \citet{shu2019understanding} calculate gradient variance by adding randomly sampled Gaussian noise on the weights of super-net, and find the optimization process is noisier and less efficient in complex search space such as stacking more cells. \citet{pourchot2020share} 
observe that, for weight sharing algorithms, architectures with a residual connection or a 3$\times3$ convolution on the first node are preferred. \citet{laube2021exploring} explore various training settings, including regularization, 
learning rate schedule and gradient clipping, and study their effects on the rank correlation of sampled single path one-shot NAS approaches~\cite{guo2020single}. These methods study how different training configurations and search spaces affect the results of weight sharing algorithms. Different from these works, we try to understand and mitigate the interference on the shared weights and improve the rank ability of super-net. 

Previous works have noticed the interference issue~\cite{bender2018understanding,guo2020single,laube2021exploring,xie2020weight}. Among them, \citet{zhang2020deeper} conduct experiments to study the interference of weight sharing using a search space of 64 architectures. By randomly sampling architectures per step for training and plotting the accuracy of previous sampled architectures, they find that current updating with the sampled child model is detrimental to other models and thus causes high variance of the rank.
In contrast, our quantitative experiment analyses reveal that the essence of the interference is different gradient directions on shared weights caused by different topologies of child models. 
Several works alleviate the interference with search space pruning. 
\citet{zhang2020deeper} propose to divide the search space according to similarities of child models, such as sorting the models lexicographically and evenly slice these models into several groups, then train the super-net on individual groups of models. Moreover, to obtain more accurate performance, they directly shrink the size of search space to one child model and fine-tune it individually. However, for a large search space, identifying similar architectures is difficult and fine-tuning each child model could incur significant computational overhead. Other methods~\cite{zhang2020does,HuLGWZWGS20Angle,xu2021taskagnostic} progressively shrink the search space by removing the unpromising architectures and operators in the training process to reduce interference. \citet{ning2021evaluating} propose several practical approaches such as operation pruning, progressively search space pruning and removing affine operations in batch normalization to reduce sharing extent. Different from these methods, inspired by our analyses, our work mitigates interference by modifying the sampling procedure and aligning inputs and outputs of shared operators. Thus, progressively shrink (pruning) methods are complementary to our methods. \citet{chu2019fairnas} also modify the sampling process by accumulating the updates over $k$ samples, chosen such that each of the $k$ operators of the super-net appears exactly once in the super-net to ensure fairness of operators updating, which increase the training costs by over $k\times$ and cannot avoid the interference between different training steps. Our work aims at quantitatively analyzing the interference and proposing effective methods to mitigate it.

\section{Analyzing Interference} \label{sec:pre_exp}
In single path one-shot NAS approaches~\cite{bender2018understanding,li2020random,guo2020single,cai2019once}, the operator shared by different child models receives different gradients that result in different optimization directions, which cause the interference among different child models. Such interference causes insufficient training and inaccurate performance evaluation, which affects the rank of child models. If we can understand what factors may influence the differences of gradients from different child models, we can gain insights for designing better solutions to mitigate the interference. Thus, in this section, we conduct a series of experiments to analyze interference.

\subsection{Analysis Setup}\label{sec:search_space_design}
The gradients are influenced by various factors including input data, weights of the super-net and selected child model. To study the gradient changes solely caused by different child models, we first train a super-net using single path one-shot NAS algorithm~\cite{guo2020single}. Then, we freeze the weights of super-net and only study the gradients caused by different child models under the same batch of training data. The super-net is organized in chain-style with $N$ layers. Each layer contains all candidate operators in $\mathcal{O}=\{o_1,\cdots,o_C\}$, where $C$ is the number of predefined candidate operators. A child model is a single path from the bottom layer to the top layer. Thus, there are $C^N$ possible child models in the super-net.
To study the interference in more challenging settings, we adopt a hybrid BERT search space including several popular candidate operators - multi-head attention~(MHA), feed-forward network~(FFN) and convolution~(CONV), and conduct experiments on the BERT pre-training task~\cite{devlin2019bert}. Specifically, to study the interference both within the same type operators and between different types of operators, we use candidate operators $\mathcal{O}=$\{MHA6, MHA8, FFN, FFN', CONV3, CONV5\}, where MHA6 is MHA with 6 heads, CONV3 is convolution with kernel size 3 and FFN' is FFN with a slightly larger inner hidden size. To exclude the influence of parameter size of different operators, we further adjust the inner hidden size of operators to ensure that they have similar parameter sizes (see Appendix~\ref{appendix:analyses_interf} for more details). We train an $N=12$ layer super-net using a batch of 1024 sentences on 32 NVIDIA P40 GPUs until 62,500 steps
. Then we freeze the super-net, and study the gradients of different child models by feeding the same batch of data (a large batch of 2048 sentences).

\begin{figure}[htb]
    \centering
    \vspace{-5pt}
    \includegraphics[width=0.4\textwidth]{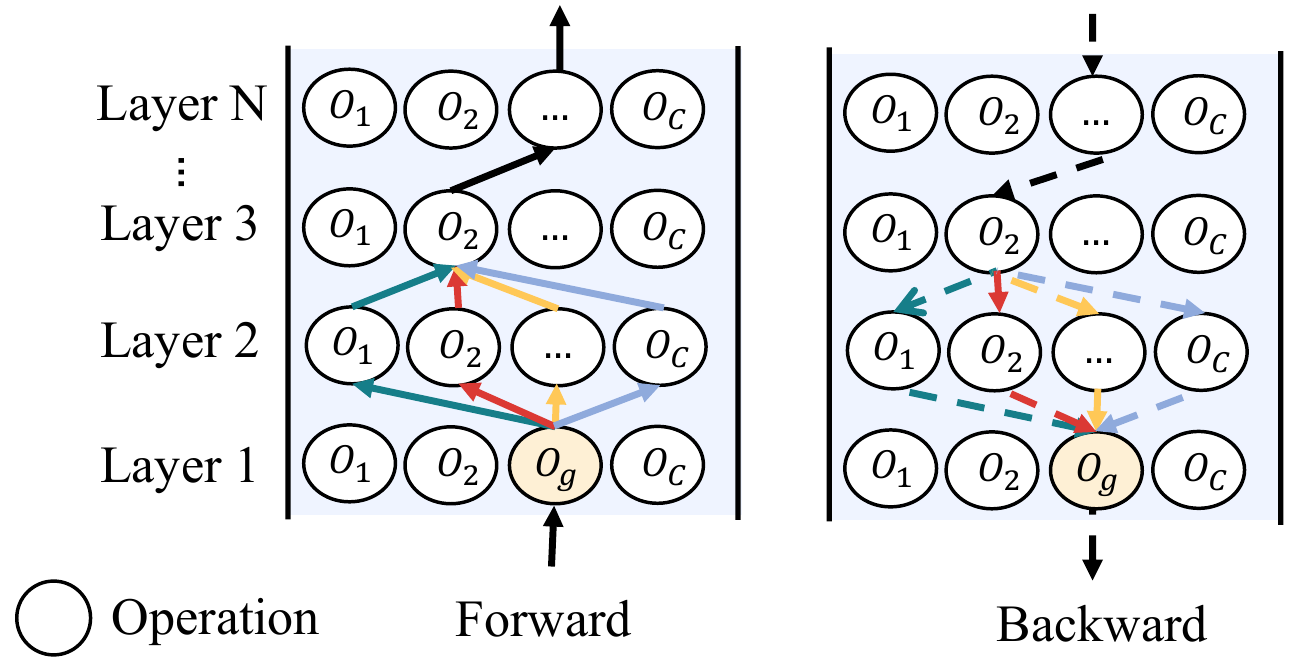}
\caption{The illustration of the forward and backward process regarding the operator $o_g$ in layer 1 that is shared by child models that differ in layer two.}
\label{fig:inteference}
\end{figure}

\subsection{Results and Analyses} \label{sec:pre_analyses_detail}
 \label{sec:inter_analyse}

\begin{figure*}[htb]
  \hspace*{0.4in}\includegraphics[width=0.9\textwidth]{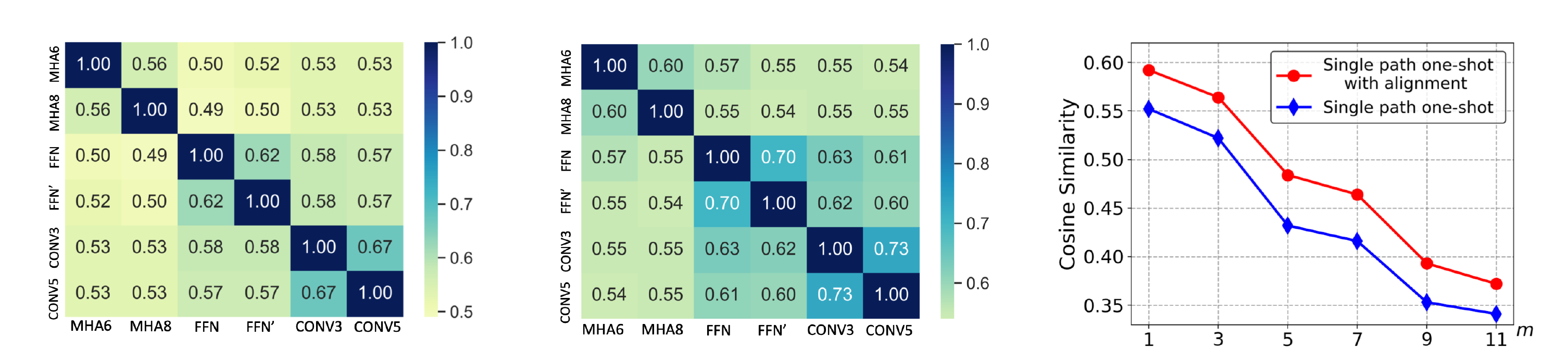} \vspace{-7mm}\\
  
  \begin{tabular}{ccc}
  \hspace*{0.5in} \tiny \begin{tabular}[c]{@{}c@{}}\tiny (a) Cosine similarity matrix of the super-net\\ \tiny trained by single path one-shot\end{tabular}    &  \hspace*{0.3in}\tiny \begin{tabular}[c]{@{}c@{}}\tiny (b) Cosine similarity matrix of the super-net\\ \tiny trained by single path one-shot with alignment\end{tabular}  & \hspace*{0.6in}\tiny (c) Gradient similarity with respect to $m$ \\
  \end{tabular}

\caption{Gradient cosine similarity. In Figures (a) and (b), the value represents the cosine similarity between two child models. For example, MHA6-FFN-0.50 in row 1 and column 3 of the similarity matrix in Figure (a) refers to the cosine similarity of the gradients on operator $o_g$ between two paths is 0.50 where the operator in layer 2 of one path is MHA6, and that in another path is FFN. For Figure (b), the super-net is obtained by training with an extra alignment training objective, and then is used to calculate gradient cosine similarity. In Figure (c), we choose child models that differ in $m$ operators and calculate the average value of the cosine similarity matrix. In each experiment, we randomly choose an $o_g$ and $C$ paths as shown in Figure~\ref{fig:inteference}. The presented results are an average of 10 repeated experiments.}
\label{fig:interference_results}
\end{figure*}

We first focus on a simple case: child models that only differ in one operator. As shown in Figure~\ref{fig:inteference}, we choose $C$ child models and they only differ in layer 2. Then we compare their gradients on $o_g$ at the first layer\footnote{In fact, the gradients on other shared operators are also different. We take $o_g$ as an example for the analyses.}. To measure the gradients similarity on $o_g$ between different child models, we flatten the gradients and calculate their cosine similarity\footnote{The cosine similarity between vector $\bm{g_i}$ and $\bm{g_j}$ is calculated by $\frac{\bm{g_i}\cdot \bm{g_j}}{||\bm{g_i}||||\bm{g_j}||}$.}. A lower cosine similarity indicates larger difference in gradients and thus more interference on the shared operators. As shown in Figure~\ref{fig:interference_results} (a), we can find 1) although different child models differ by only one operator, the gradient cosine similarity is still not high, indicating that different child models can actually lead to very different gradient directions on their shared operators; 2) the same type of operators have less gradient interference (larger cosine similarity) than different types of operators, demonstrating that the interference in hybrid search space tends to be more serious than those with the same type operators.

For the hybrid search space, although all child models are trained under the same task, there is no guarantee that the operator should learn the same transformation on different child models. In other words, the inputs and outputs of the operator differ when it locates on different child models. It motivates us to explicitly force the inputs and outputs of the operators to be similar to reduce such interference. To this end, we re-train a super-net. At each training step, we first randomly choose $C$ child models and calculate their average inputs and outputs in each layer. Then we randomly sample a child model for training and add an extra alignment training objective by calculating MSE loss between its inputs and outputs and the average ones in a layer-by-layer manner~(see detailed formulation in Appendix~\ref{appendix:analyses_interf}). Finally, we freeze the super-net and re-calculate the gradient cosine similarity as aforementioned. As shown in Figure~\ref{fig:interference_results} (b), we can observe that \emph{by aligning the inputs and outputs of the shared operators to be similar to the average inputs and outputs, the gradient interference can be reduced}.

We further extend the analyses to a more general case: child models that differ in $m$ operators. Specifically, we choose $C$ child models that differ from the second layer to the $(m+1)$-th layers and obtain the cosine similarity matrix on $o_g$ similar to Figure~\ref{fig:interference_results} (a). Then we calculate the average of the cosine similarity matrix to represent the average interference by varying $m$ operators. The results are shown in Figure~\ref{fig:interference_results} (c). As we can see, \emph{the interference on a shared operator between two child models is positively correlated with the number of different operators between them}. It demonstrates that randomly sampling child models in the training process in a complex search space could cause serious interference since their architecture topologies may differ a lot.

Although the analyses are performed under conditions of freezing the super-net and feeding the same batch of data, our findings about which factors influence the interference are general, and can help develop new methods to mitigate interference during training, as described in the next section.


\section{Mitigating Interference}
Inspired by the observations in Sec.~\ref{sec:pre_exp}, we propose two methods: MAGIC-T and MAGIC-A, to mitigate the interference.  

\paragraph{\underline{M}itig\underline{A}tin\underline{G} \underline{I}nTerferen\underline{C}e from the pespective of \underline{T}opological environment~(MAGIC-T). }
According to our analyses in Sec.~\ref{sec:inter_analyse}, more topological differences between two child models can cause more interference on the shared operators. 
Given a super-net with $N$ layers and $C$ candidate operators, previous single path NAS approaches~\cite{bender2018understanding,chu2019fairnas,li2020random,guo2020single} change $\frac{N(C-1)}{C}$ operators between $\alpha_t$ and $\alpha_{t-1}$ on average, where $\alpha_t$ is the sampled child model at step $t$, because there are $N$ layers (operators) and the probability of each operator in $\alpha_t$ being different from the one in $\alpha_{t-1}$ is $\frac{C-1}{C}$.  Modern algorithms usually employ a huge search space with dozens of layers~\cite{pham2018efficient,liu2018darts,cai2019once,chu2019fairnas}. For such a search space, two child models sampled between adjacent sampling steps differ a lot and the gradients of shared operators interfere with each other, 
which is detrimental to making progress in training. 
To mitigate such interference, we propose MAGIC-T to gradually change the topological environment. Specifically, at each training step, MAGIC-T samples a child model $\alpha_t$ by randomly substituting one operator ($k=1$) in the child model $\alpha_{t-1}$ sampled at the last step with another operator, and applies the forward and backward computation using $\alpha_t$ for weights updating. 

Although MAGIC-T substitutes only one operator in one iteration, it does not only sample locally around 
the initial child model $\alpha_0$. Indeed, as $t$ increases, MAGIC-T can sample diverse architectures, approaching to the uniform sampling over the entire search space.  This can bee shown by viewing MAGIC-T performing a random walk on the {\em  architecture graph} $\calG$, in which each architecture corresponds to a node and two nodes are connected by an edge if and only if they differ in only one operator. It is easy to see that $\calG$ consists of $C^N$ nodes (recall that $N$ is the number of layers and $C$ is the number of operators) and the (shortest) distance between any two nodes is at most $N$. 
Suppose the random walk $\{X_t\}_t$ starts at node $X_0=\alpha_0$, and denote the node distribution at time $t$ as $\pi_t$ ($\pi_t$ is a $C^N$ dimensional vector and $\pi_t(v)=\Pr[X_t=v]$).
Let $\pi$ be the uniform distribution over the nodes of $\calG$ (which is the stationary distribution).
By standard convergence theory of random walk in Markov chain~\cite{levin2017markov}\footnote{
This can be proved using the standard coupling argument. See the example
of random walk on the hypercube in Sec.~5.3 \cite{levin2017markov}. Although our graph $\calG$
is slightly more general than hypercube, the same argument applies.
}
, one can show that
the total variational distance between $\pi_t$ and $\pi$
$$
d_{TV}(\pi_t, \pi) \leq \exp(-t/N-\ln N).
$$
Hence, if we want $d_{TV}(\pi_t, \pi)\leq \epsilon$, we only need $t\geq N\ln N+N\log 1/\epsilon$. Since $t$ is sufficiently large (e.g., 250,000 steps for BERT), the variational distance between the node distribution of MAGIC-T and uniform sampling is sufficiently small. Thus, MAGIC-T can sample diverse architectures from the whole search space like uniform sampling.

\paragraph{\underline{M}itig\underline{A}tin\underline{G} \underline{I}nTerferen\underline{C}e from the perspective of inputs and outputs \underline{A}lignment (MAGIC-A).}

In Sec.~\ref{sec:inter_analyse}, 
we find that using the average inputs and outputs of shared operators for alignment can reduce the interference. However, it increases the computational cost by a factor of $C$, and limits the flexibility of search algorithms. Thus, we directly pick a top-performing anchor child model from the search space to align other child models. The anchor model can be replaced when the performance of another child model outperforms it. Formally, the inputs and outputs alignment loss between sampled child model $\bm{\alpha_t}$ and the anchor model $\bm{\alpha^l}$ is defined as
\begin{equation}\label{eqn:align_loss}
    \mathcal{L}_{align}(\mathbf{H}^{(\bm{\alpha^l)}}, \mathbf{H}^{(\bm{\alpha_t})})=\sum_{n=1}^N \text{MSE}(\mathbf{H}_n^{(\bm{\alpha^l)}}, \mathbf{H}_n^{(\bm{\alpha_t})}),
\end{equation}
where $\mathbf{H}_n$ is the outputs of $n$-th layer (the inputs of ($n+1$)-th layer),
and $N$ is the number of layers. Then the training objective of a sampled child model $\bm{\alpha_t}$ is
\begin{equation}\label{eqn:pred_loss}
    \mathcal{L} =  \mathcal{L}_{pred} + \lambda \mathcal{L}_{align},
\end{equation}
where $\mathcal{L}_{pred}$ is the supervised loss between predictions and ground truth (e.g., cross entropy loss in masked language modeling), $\lambda$ is the scaling parameter that controls the weight of alignment loss.  In practice, we can adopt block-wise alignment where each block contains a few layers to avoid over-regularization. The training procedure of MAGIC-A at each step is as follows:

\begin{itemize}[leftmargin=*]
    \item Obtain a batch of data and an anchor child model $\bm{\alpha}^l$, and randomly sample a child model $\bm{\alpha_t}$,
    \item Calculate the loss according to Eq.~(\ref{eqn:pred_loss}) and update the weights of $\bm{\alpha_t}$,
    \item Replace $\bm{\alpha}^l$ with $\bm{\alpha_t}$ if $\text{Val}(\bm{\alpha}_t)>\text{Val}(\bm{\alpha}^l)$,
\end{itemize}
where $\text{Val}(\cdot)$ is the accuracy obtained from the dev set. 
Maintaining a top-performing anchor model is similar to prioritized paths introduced by \citet{peng2020cream}. However, our anchor model is used to align inputs and outputs of shared operators and reduce interference rather than distillation on the final layer to boost the training of child models.

\section{Experiments and Results}
\subsection{Setup} \label{exp:setup}
\paragraph{Search space and super-net training.}  We adopt a chain-styled super-net with candidate operators $\mathcal{O}=$\{MHA12, FFN,  CONV3, CONV5\}. Following BERT~\citep{devlin2019bert}, we train the super-net and discover architectures using BookCorpus plus English Wikipedia~(16GB in total). The super-net is trained with the batch size of 1024 sentences for 250,000 steps (same computational resources as BERT$_{\rm base}$). We adopt commonly-used SPOS~\cite{guo2020single} as the baseline method, which randomly samples a child model for training at each step. For MAGIC-T, unless otherwise mentioned, it gradually changes $k=1$ operator to reduce the interference. For MAGIC-A, we train the super-net with masked language modeling loss in the first three epochs for a warm-start and then add the alignment loss with $\lambda=0.5$ in every four layers to avoid over-regularization. Since MAGIC-T and MAGIC-A are independent methods proposed from two different perspectives, they can also be combined (MAGIC-AT) to collaboratively mitigate the interference. See Appendix~\ref{appendix:arches_and_configs} for detailed configurations.

\paragraph{Architecture search and re-training.} Previous single path one-shot methods usually adopt progressively shrinking~\cite{xu2021taskagnostic,HuLGWZWGS20Angle} or evolution algorithms~\cite{guo2020single,cai2019once} to search effective models. Since progressively shrinking can allocate more computational resources to promising architectures and speed up the search process, we adopt it by deleting five unpromising operators~\cite{xu2021taskagnostic} at the end of each epoch until only one child model is left in the search space. Note that we do not adopt progressively shrinking when analyzing the rank correlation in Sec.~\ref{sec:nas_bert_analyse}. To evaluate the standalone performance of child models, we re-train them from scratch for 125,000 steps with batch size 2048 on 32 NVIDIA P40 GPUs and keep other configurations the same as those of super-net. \footnote{The total computational resources (125,000*2048 sentences of 16GB corpus) is exactly the same as BERT$_{\rm base}$~\cite{devlin2019bert}, MPNet$_{\rm base}$~\cite{Song0QLL20} and ELECTRA$_{\rm base}$~\cite{ClarkLLM20} for a fair comparison.} We train RoBERTa$_{\rm base}$~\cite{liu2019roberta} following its original configurations while using the same computational resources as BERT$_{\rm base}$. 

\paragraph{Evaluation.}
We evaluate performance by fine-tuning pre-trained models on GLUE benchmark~\cite{WangSMHLB19} 
and follow hyper-parameters of \citet{liu2019roberta} for fine-tuning.


\subsection{Analyses on the Rank Correlation}\label{sec:nas_bert_analyse}
\paragraph{Comparison of Kendall rank correlation.}
We first perform correlation analysis to evaluate whether our method MAGIC can improve the rank of child models by mitigating interference. First, we uniformly sample 60 child models, train them on the pre-training tasks and obtain their ground-truth performance on the downstream MNLI task~\citep{williams2018mnli}, which is widely used to indicate the performance of the NLP pre-training models~\cite{liu2019roberta,Song0QLL20,Unilm2019}. Then, given a super-net, we can calculate the Kendall rank correlation coefficient (Kendall's Tau)~\cite{kendall1938new} between their negative weight sharing validation loss and ground-truth performance. As shown in Table~\ref{table:bert_correlation}, compared with the baseline, MAGIC-T improves the rank correlation from 0.36 to 0.49, which shows the benefits of gradual modification to mitigate the interference. Furthermore, combining MAGIC-T and MAGIC-A achieves an even higher rank correlation of 0.57. It shows that MAGIC-T and MAGIC-A, proposed from two different perspectives, can collaboratively reduce interference to further improve the rank correlation.

\begin{table}[htbp]
\centering
\caption{Kendall's Tau on the BERT search space.}
\label{table:bert_correlation}
\scalebox{0.7}{
\begin{tabular}{l|ccccc}
\toprule
\multirow{2}{*}{Method}      & SPOS & \multirow{2}{*}{MAGIC-T} & \multirow{2}{*}{MAGIC-A} & \multirow{2}{*}{MAGIC-AT} \\
& \cite{guo2020single} & & & \\\midrule
Kendall's Tau &    0.36  &  0.49  &      0.44    &   \textbf{0.57}   \\\bottomrule
\end{tabular}}
\end{table}

\begin{figure*}[htb]
  \hspace*{0.30in}\includegraphics[width=0.9\textwidth]{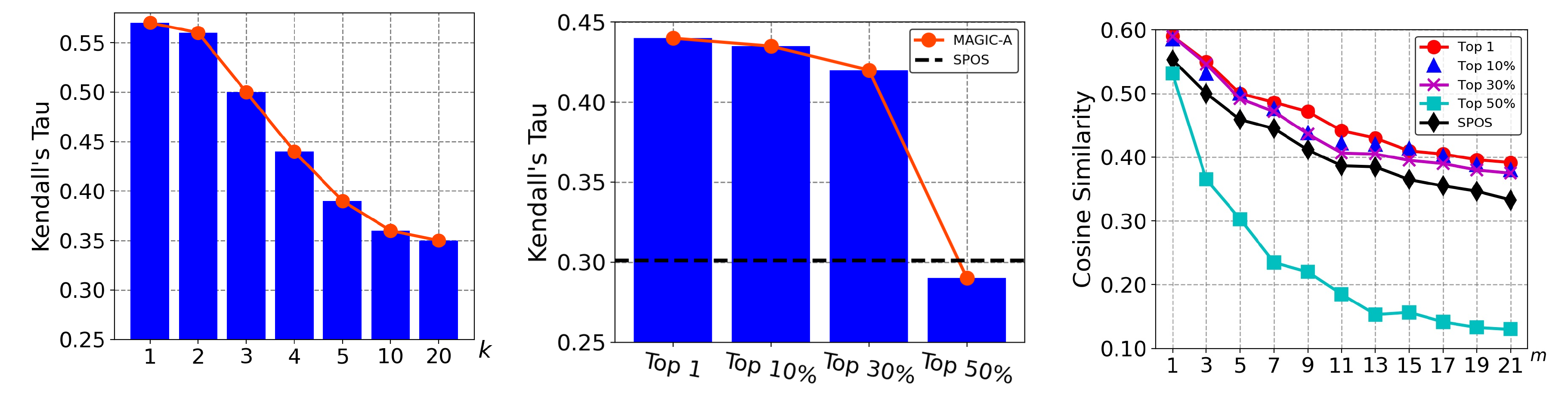} \vspace{-3mm}\\
  \begin{tabular}{ccc}
  \hspace*{0.60in} \scriptsize (a) Study MAGIC-T by varying $k$.  &  \hspace*{0.6in}\scriptsize(b) Study of MAGIC-A. & \hspace*{0.6in}\scriptsize (c) Gradient cosine similarity of MAGIC-A. \\
  \end{tabular}
\caption{Analyses of MAGIC-T and MAGIC-A.}
\label{fig:methods_analysis}
\end{figure*}

\begin{table*}[!t]
\centering
\caption{Results of pre-training models on GLUE benchmark. The test set results are obtained from the official GLUE leaderboard. All models are trained under the same computational resources as BERT$_{\rm base}$  for a fair comparison. ``E-'' refers to training in the ELECTRA style. Following~\cite{devlin2019bert}, Spearman correlation is reported for STS-B, Matthews correlation is reported for CoLA and accuracy is reported for other tasks.}
\label{main_resuls}
\scalebox{0.75}{
\begin{tabular}{l|l|l|llllllll|l}
\toprule
Model & Params & FLOPs & MNLI & QQP  & QNLI & CoLA & SST-2 & STS-B & RTE  & MRPC & AVG    \\
\midrule
\multicolumn{11}{l}{\emph{dev set}} \\\midrule
BERT$_{\rm base}$~\cite{devlin2019bert}      & 110M  & 2.9e10  & 84.4   & 89.9 & 88.4 & 54.3 & 92.7  & 88.9  & 71.1 & 86.7  &  82.1  \\
RoBERTa$_{\rm base}$~\cite{liu2019roberta}   & 125M  & 3.3e10 &  85.3  & 91.1 & 91.1 & 61.0 & 92.7  & 90.0 & 77.5 & 87.9  & 84.6   \\
ELECTRA$_{\rm base}$~\cite{ClarkLLM20} & 110M   & 2.9e10 & -   & - & - & - & -  & -  & - & - & 85.1\\
MPNet$_{\rm base}$~\cite{Song0QLL20}   & 110M   & 2.9e10 & 85.2   & - & - & -    & 93.4  & -     & - & - & -\\
SPOS~\cite{guo2020single}   & 114M   & 3.3e10 &  84.7  & 91.4 & 91.4 &  59.6   & 92.1  & 89.7 & 80.9 & 86.3 & 84.4\\\midrule
MAGIC-AT    & 113M     & 3.3e10     &  85.6  & 91.3 &  91.8 & 61.1 &  \textbf{93.5} & 90.3  & 80.9 & \textbf{90.9} & 85.7\\
E-MAGIC-AT     & 110M     & 2.9e10 &  \textbf{86.3}  & \textbf{91.7} & \textbf{92.5}  & \textbf{65.8} &  92.5 & \textbf{91.0}  & \textbf{84.0} & 89.7 & \textbf{86.7}\\
\bottomrule
\toprule
\multicolumn{10}{l}{\emph{test set}} \\
\midrule
BERT$_{\rm base}$~\cite{devlin2019bert} & 110M    & 2.9e10&  84.6 & 89.2  & 90.5 & 52.1 & 93.5  & 85.8  & 66.4 & 84.8 &  80.9    \\
RoBERTa$_{\rm base}$~\cite{liu2019roberta} & 125M  & 3.3e10  &  84.8  & 89.0 & 91.7 & 57.1 & 93.3 & 88.0   & 74.1 & 84.1 & 82.8    \\
ELECTRA$_{\rm base}$~\cite{ClarkLLM20} & 110M  & 2.9e10 &  85.8  & 89.1 & \textbf{92.7} &  59.7 & 93.4 &  87.7  & 73.1    &  86.7  & 83.5\\
SPOS~\cite{guo2020single}   & 114M   & 3.3e10 &  84.3  & 88.6 & 91.0 &  56.1   &  92.8 &  88.1     & 74.9 & 83.4 & 82.4\\\midrule
MAGIC-AT       & 113M & 3.3e10 &  84.9  & 89.1 &  92.0 & 57.0 &  \textbf{94.1} & \textbf{87.8}  & 77.4 & 85.2 & 83.4\\
E-MAGIC-AT     & 110M & 2.9e10 &  \textbf{85.9}  & \textbf{89.6} & 92.4  & \textbf{60.3} &  93.4 & 87.3  & \textbf{80.4} & \textbf{87.4} & \textbf{84.6}\\
\bottomrule
\end{tabular}}
\end{table*}

\paragraph{Analyses of MAGIC-T.}
MAGIC-T gradually modifies $k=1$ operators at each training step to reduce the interference. We further analyze MAGIC-T by varying $k$ and measuring the rank correlation. According to analyses in Sec.~\ref{sec:pre_analyses_detail}, more topological changes (larger $k$) can lead to more serious interference on the shared operators. As shown in Figure~\ref{fig:methods_analysis} (a), it is clear to observe that larger $k$ causes worse rank correlation, which demonstrates that the interference is detrimental to the super-net training, and our proposed MAGIC-A can efficiently mitigate the interference and improve the rank ability of the super-net.

\paragraph{Analyses of MAGIC-A.}
MAGIC-A selects a top-performing anchor model to align inputs and outputs of other child models to be similar to reduce the interference. Besides a top-performing one, we further analyze the method by choosing different anchor models. Specifically, we evaluate $10,000$ child models at the end of every epoch and select a top $p\%$ child model as the anchor model. However, different from the top-performing model, the exact top $p\%$ model varies dramatically, which results in selecting different anchor models in different epochs. Using such anchor models to align other models in each epoch can lead to unstable optimization of the super-net. For stable training, we substitute the anchor model when the previously selected anchor model in the last epoch escapes from the range of top $p\pm r$\% models ($r=10$) in the current epoch. We train different super-nets with different $p$ and 
present the ranking correlation in Figure~\ref{fig:methods_analysis} (b). The cosine similarity results, as introduced in Sec.~\ref{sec:pre_analyses_detail}, are presented in Figure~\ref{fig:methods_analysis} (c).
We can observe that a proper anchor model that ranks within the top 30\% is enough to align the inputs and outputs and mitigate interference, and a better anchor model can obtain slightly better results. However, choosing a model that performs poorly as the anchor model has a negative effect on super-net training. Since the optimization goal is $\mathcal{L}_{pred} + \lambda \mathcal{L}_{align}$, alignment regularization guided by a  bad child model may deviate with the training target $\mathcal{L}_{pred}$. 

In the following sections, we demonstrate the effectiveness and generality of MAGIC-AT on the BERT pre-training, compression task and image classification tasks.

\subsection{Searching Effective BERT Models} \label{sec:110M_model}
\paragraph{Results on GLUE benchmark.}
To show the generality of the architecture discovered by our MAGIC-AT, we train it with two different settings: 1) masked language modeling (MLM)~\cite{devlin2019bert} proposed in BERT~\cite{devlin2019bert} and 2) replace token detection (RTE) proposed in ELECTRA~\cite{ClarkLLM20}. The discovered architecture is shown in Figure~6 and the experimental results are shown in Table~\ref{main_resuls}. Using the MLM object, our model consistently outperforms other models by a large margin on both the dev and the test set. Compared with RoBERTa, our model achieves 1.1 and 0.6 higher points on the dev and the test set respectively. Using replace token detection objective following ELECTRA, our model is superior to ELECTRA by 1.6 and 1.1 points on the dev and the test set respectively.

\begin{table*}[tbh]
\centering
\caption{Comparison of compression methods in the commonly-used 60M model size. ``*'' means using data augmentation. }
\label{table:compression}
\scalebox{0.80}{
\begin{tabular}{l|l|llllllll|l}
\toprule
Model & Params  & MNLI & QQP  & QNLI & CoLA & SST-2 & STS-B & RTE  & MRPC & AVG    \\
\midrule
\multicolumn{11}{l}{\emph{dev set}} \\\midrule
DistilBERT~\cite{sanh2019distilbert}      & 66M  & 82.2   & 88.5 & 89.2 & 51.3 & 91.3  & 86.9  & 59.9 & 87.5  &  79.6  \\
MiniLM~\cite{wang2020minilm}           & 66M   & 84.0   & 91.0 & 91.0 & 49.2 & 92.0  & -     & 71.5 & 88.4  & -    \\
BERT-of-Theseus~\cite{xu2020bert} & 66M    & 82.3   & 89.6 & 89.5 & 51.1 & 91.5  & 88.7  & 68.2 & - & -\\
PD-BERT~\cite{turc2019well}          & 66M    & 82.5   & 90.7 & 89.4 & -    & 91.1  & -     & 66.7 & 84.9 & -\\
DynaBERT*~\cite{hou2020dynabert}        & 60M    & 84.2   & 91.2 & 91.5 & 56.8 & 92.7  & 89.2  & 72.2 & 84.1 & 82.7\\
NAS-BERT~\cite{xu2021taskagnostic} & 60M  & 84.1 & 91.0 & 91.3  & 58.1   & 92.1 & 89.4 & 79.2  &   88.5 & 84.2 \\
SPOS~\cite{guo2020single}       & 60M    &  84.0  & 90.7  & \textbf{91.1} & 57.1 & 91.6 & 88.2 & 75.9 & 86.5 & 83.1\\
\midrule
MAGIC-AT            & 60M    &  84.5  & 90.9 &   91.1 &  61.8 &  92.8 & 89.0  & 78.9 & 89.2 & \textbf{84.8}\\
\bottomrule
\toprule
\multicolumn{10}{l}{\emph{test set}} \\
\midrule
BERT-of-Theseus~\cite{xu2020bert} & 66M    &  82.4 & 89.3 & 89.6 & 47.8 & 92.2  & 84.1  & 66.2 & 83.2 & 79.4     \\
PD-BERT~\cite{turc2019well}           & 66M    &  82.8  & 88.5 & 88.9 & -    & 91.8  & -     & 65.3 & 81.7 & -     \\
BERT-PKD~\cite{sun2019patient}         & 66M    &  81.5  & 88.9 & 89.0 &  - & 92.0 &  -  & 65.5     &  79.9  & -\\
TinyBERT*~\cite{jiao2019tinybert}         & 66M    &  84.6  & 89.1 & 90.4 &  51.1 & 93.1 & 83.7  & 70.0 & 82.6  & 80.6\\
NAS-BERT~\cite{xu2021taskagnostic}             & 60M    & 83.5   & 88.9 & 90.9 & 48.4 & 92.9  & 86.1  & 73.7 & 84.5 & 81.1\\
SPOS~\cite{guo2020single}       & 60M    &  83.5  & 88.5  & \textbf{90.6} & 52.4 & 91.7 & 86.5 & 74.2 & 83.6 & 81.4\\
\midrule
MAGIC-AT & 60M &  84.2  & 88.8 & 90.6 & 53.6 & 92.1 & 86.8 & 75.6 & 84.3 & \textbf{82.0}\\
\bottomrule
\end{tabular}}
\end{table*}

\begin{table}[htb]
\centering
\caption{Results on the dev set of SQuAD datasets.}
\label{table:SquAD_results}
\scalebox{0.75}{
\begin{tabular}{l|rr | rr}
\toprule
\multicolumn{1}{c|}{\multirow{2}{*}{Model}} & \multicolumn{2}{c|}{SQuAD v1.1}                   & \multicolumn{2}{c}{SQuAD v2.0} \\
\multicolumn{1}{c|}{}   & \multicolumn{1}{c}{EM} & \multicolumn{1}{c|}{F1} & \multicolumn{1}{c}{EM} & \multicolumn{1}{c}{F1}   \\
\midrule
BERT$_{\rm base}$~\cite{devlin2019bert}                                                                         & 80.5           & 88.5          & -              & -             \\
RoBERTa$_{\rm base}$~\cite{liu2019roberta}                                                                             & 82.1           & 89.3          & 74.9           & 78.2          \\
ELECTRA$_{\rm base}$~\cite{ClarkLLM20}                                                                             & 84.5              & 90.8             & 80.5              & 83.3          \\\midrule
MAGIC-AT                                                                          &       82.7     &  90.0         &      76.6     &    80.4    \\
E-MAGIC-AT       &      \textbf{84.8}      &    \textbf{91.4}      &     \textbf{80.8}      &     \textbf{83.9}    \\
\bottomrule
\end{tabular}
}
\end{table}

\paragraph{Results on SQuAD datasets.}
We further evaluate the generalizability of our searched architecture by fine-tuning it to reading comprehension tasks SQuAD v1.1~\citep{rajpurkar2016squad} and SQuAD v2.0~\citep{rajpurkar2018know}. We adopt standard evaluation metrics of Exact-Match (EM) and F1 scores following~\cite{devlin2019bert,liu2019roberta,ClarkLLM20}. 
The results are shown in Table~\ref{table:SquAD_results}. Compared with RoBERTa, our model achieves 0.6 and 0.7 higher points on EM and F1 for SQuAD v1.1, 1.7 and 2.2 higher points on EM and F1 for SQuAD v2.0 while using fewer parameters. Furthermore, our model also surpasses the ELECTRA using replace token detection objective.

\subsection{Searching Compressed BERT Models}
We further validate the generalizability of our algorithms for BERT model compression. Previous works on BERT compression aim to design novel architectures to explore the potential of different architectures~\cite{xu2021taskagnostic} or advanced distillation methods~\citep{sanh2019distilbert,wang2020minilm,turc2019well,hou2020dynabert} to efficiently learn knowledge from the teacher model. We build a super-net as described in Sec.~\ref{exp:setup} but reduce the hidden size of candidate operators from 768 to 512. We re-run MAGIC-AT on this search space and evaluate the architecture found by our algorithm for model compression. Following NAS-BERT~\citep{xu2021taskagnostic}, we apply knowledge distillation on two stages to train the discovered model (i.e., pre-training and fine-tuning) and do not add extra techniques such as attention matrix distillation~\citep{jiao2019tinybert,hou2020dynabert}. We adopt the teacher model used in NAS-BERT~\cite{xu2021taskagnostic} to ensure that the improvement is caused by a superior architecture. Other training hyper-parameters are the same as those in Sec.~\ref{sec:110M_model}. The discovered architecture and detailed configurations are presented in Appendix~\ref{appendix:arches_and_configs}. As shown in Table~\ref{table:compression}, our model achieves better performance compared to all previous approaches. This again shows the general effectiveness of our algorithms.

\begin{table}[htb]
\centering
\caption{Comparison of models on ImageNet. }
\vspace{-6pt}
\label{table:imagenet_results}
\scalebox{0.70}{
\begin{tabular}{l|cccc}
\toprule
Model & Top1/Top5 Err. & Params & FLOPS \\
\midrule
MobileNetV2~\cite{sandler2018mobilenetv2} & 25.3/- & 6.9M & 585M \\
ShuffleNetV2~\cite{zhang2018shufflenet} & 25.1/- & $\sim$5M & 591M \\
\midrule
DARTS~\cite{liu2018darts} & 26.9/9.0 & 4.9M & 595M \\
PC-DARTS~\cite{xu2019pc} & 24.2/7.3 & 5.3M & 597M \\
CARS~\cite{yang2020cars} & 24.8/7.5 & 5.1M & 591M \\
PC-NAS~\cite{li2020improving} & 23.9/- & 5.1M & - \\
EnTranNAS-DST~\cite{Yang_2021_CVPR} & 23.8/7.0 & 5.2M & 594M \\\midrule
\multicolumn{3}{l}{\emph{Models searched on the MobileNetV2 search space}} \\\midrule
NAO~\cite{luo2018neural} & 24.5/7.8 & 6.5M & 590M \\
LaNAS~\cite{wang2021sample} & 25.0/7.7 & 5.1M & 570M \\
BN-NAS~\cite{Chen_2021_ICCV} & 24.3/- & 4.4M & 470M \\
ProxelessNAS~\cite{cai2018proxylessnas} & 24.0/7.1 & 5.8M & 595M \\
RLNAS~\cite{Zhang_2021_CVPR} & 24.4/7.4 & 5.3M & 473M \\
SemiNAS~\cite{luo2020semi} & 23.5/6.8 & 6.3M & 599M \\
\midrule
MAGIC-AT & \textbf{23.2/6.7} & 6.0M & 598M \\
\bottomrule
\end{tabular}
}
\vspace{-15pt}
\end{table}

\subsection{Searching on ImageNet}
We use a MobileNet-v2~\cite{sandler2018mobilenetv2} based search search space following ProxylessNAS~\cite{cai2018proxylessnas}, which excludes squeeze-and-excitation block~\cite{hu2018squeeze}. Candidate operations include inverted bottleneck convolution~\cite{sandler2018mobilenetv2} with various kernel sizes \{3, 5, 7\}, expansion ratios \{3, 6\} and zero-out layer. For super-net training, we use the SGD optimizer with an initial
learning rate of 0.4 and a cosine learning rate, and train the super-net on 8 V100 GPUs for 150 epochs with a batch size of 512. For stand-alone model training, to be consistent with the previous works, we follow the same strategy as ProxylessNAS\footnote{\url{https://github.com/mit-han-lab/proxylessnas}} and do not employ tricks like cutout~\cite{cutout1708-04552} or mixup~\cite{zhang2018mixup}. We mainly compare MAGIC-AT to works with the same search space and present the results in Table~\ref{table:imagenet_results}. MAGIC-AT achieves 23.2\% top-1 test error rate on ImageNet under the 600M FLOPS constraint, which outperforms baseline NAS works. The discovered architecture is depicted in 
Figure~\ref{fig:arch_imagenet}.

\section{Conclusion}\label{sec:conclusion}
In this paper, we quantitatively analyze the causes of interference on shared weights in neural architecture search and develop two approaches: MAGIC-T and MAGIC-A, to mitigate it. The proposed methods can improve the rank correlation of the super-net and can search efficient architectures. Experiments on various natural language tasks and ImageNet task demonstrate the effectiveness of our methods. We hope our quantitative analyses and approaches to mitigate interference can help the NAS community to better understand and resolve the interference issue.

\section{Acknowledgements}
Jin Xu and Jian Li are supported in part by the National Natural Science Foundation of China Grant 62161146004, Turing AI Institute of Nanjing and Xi'an Institute for Interdisciplinary Information Core Technology. 

\nocite{langley00}
\bibliography{main}
\bibliographystyle{icml2022}

\clearpage
\appendix
\section{Configurations and More Analyses of Interference}\label{appendix:analyses_interf}
In this section, we present detailed configurations of results presented in Sec.~\ref{sec:pre_exp} and more analyses of interference. We first describe the structure of operators in Sec.~\ref{setup}. Then we present how to align the inputs and outputs of shared operators with the average inputs and outputs in Sec.~\ref{sec:align}. Next, to give readers a better understanding, we depict how we extend the analyses to a more general case: child models that differ in $m$ operators in Sect.~\ref{sec:vary_m}. Finally, we discuss more results of interference when $o_g$ locates in different layers in Sec.~\ref{sec:o_g}.

\subsection{Candidate Operators}\label{setup}
\begin{figure*}[htb]
    \centering
    \hspace*{0.1in}\includegraphics[width=0.90\textwidth]{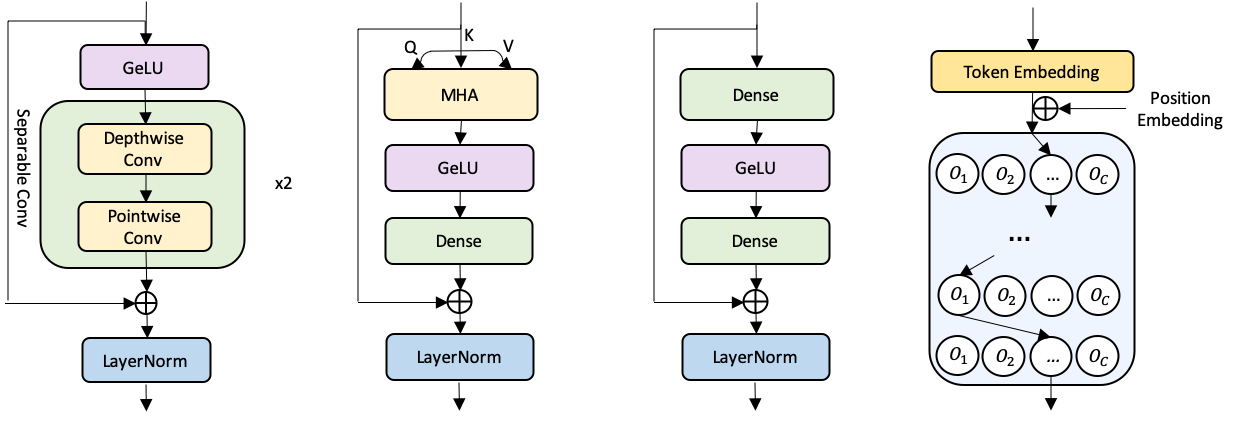}\\
  
  \begin{tabular}{cccc}
  \hspace*{0in}  (a) Conv  &  \hspace*{0.6in}(b) MHA & \hspace*{0.6in} (c) FFN  & \hspace*{0.4in} (d) Super-net\\
  \end{tabular}
\caption{The architecture of operators and the super-net for analyzing interference.}
\label{fig:op_architecture}
\vspace{-2mm}
\end{figure*}

We adopt a hybrid BERT search space including popular and different types of candidate operators: multi-head attention~(MHA), feed-forward network~(FFN) and convolution~(CONV). The structures of FFN and MHA follow those in BERT~\cite{devlin2019bert} as shown in Figure~\ref{fig:op_architecture} (b) and (c). We adopt separable convolution~\cite{sep17}, as shown in Figure~\ref{fig:op_architecture} (a), since 1) its effectiveness in natural language processing tasks have been demonstrated by previous work~\citep{kaiser2018depthwise,karatzoglou2020applying,xu2021taskagnostic} and 2) its number of parameters does not change much by varying the kernel size, which allows us to conveniently obtain different CONV operators with the similar size of parameters. To exclude the influence of parameter size of different type operators, we further adjust the inner hidden size of operators to ensure that they have a similar number of parameters as shown in Table~\ref{table:op}. 

\begin{table}[htb]
\centering
\vspace{-2mm}
\caption{Configurations of different operators used for analyzing interference.}
\label{table:op}
\scalebox{0.675}{
\begin{tabular}{l|ccc}
\toprule
         & \multicolumn{3}{c}{Candidate Operators}                 \\\midrule
$O$       & \multicolumn{3}{c}{\{MHA6, MHA8, FFN, FFN', CONV3, CONV5\}} 
 \\\midrule\midrule
Operator & Hidden Size   & Parameters  & Configurations                  \\\midrule
MHA 6    & 768           & 1.18M       & Heads 6, QKV hidden 384  \\
MHA 8    & 768           & 1.18M       & Heads 8, QKV hidden 384  \\
FFN      & 768           & 1.18M       & Inner hidden 768         \\
FFN'     & 768           & 1.28M       & Inner hidden  832        \\
CONV3    & 768           & 1.18M       & Kernel 3                 \\
CONV5    & 768           & 1.19M       & Kernel 5                 \\\bottomrule
\end{tabular}}
\vspace{-14pt}
\end{table}

\subsection{Aligning the Inputs and Outputs with the Average Inputs and Outputs} \label{sec:align}
To explicitly align the inputs and outputs of the shared operators to be similar to the average inputs and outputs, we should first enumerate all possible child models in the search space and obtains their inputs and outputs in every layer. However, it can incur significant computational overhead. Thus, we only randomly choose $C$ child models and obtain their average inputs and outputs as follows:
\begin{equation}\label{eqn:align_loss_avg}
    \mathbf{H}_n^{\text{avg}}=\frac{1}{C}\sum_{i=1}^C \mathbf{H}_n^{(\bm{\alpha^i)}},n=1,2,\cdots,N 
\end{equation}
where $\mathbf{H}_n^{(\bm{\alpha_i)}}$ is the outputs of $n$-th layer (the inputs of ($n+1$)-th layer) of child model $\bm{\alpha^i}$, $\mathbf{H}_n^{\text{avg}}$ is the average outputs of $C$ child models in $n$-th layer and $N$ is the number of layers. Then, the alignments loss is defined as 
\begin{equation}\label{eqn:align_loss}
    \mathcal{L}_{align}(\mathbf{H}_n^{\text{avg}}, \mathbf{H}^{(\bm{\alpha})})=\sum_{n=1}^N \text{MSE}(\mathbf{H}_n^{\text{avg}}, \mathbf{H}_n^{(\bm{\alpha})}),
\end{equation}
where $\alpha$ is a sampled child model for optimization and $\text{MSE}$ is mean square error loss. The training objective of a child model $\alpha$ is
\begin{equation}\label{eqn:pred_loss}
    \mathcal{L} =  \mathcal{L}_{pred} + \lambda \mathcal{L}_{align},
\end{equation}
where $\mathcal{L}_{pred}$ is the supervised loss between the predictions and the ground truth (e.g., cross entropy loss in masked language modeling), $\lambda$ is the scaling parameter that controls the weight of the alignment loss and we adopt $\lambda=0.5$ in our experiments.

\subsection{The Setting about Interference of the Operator Shared by Child Models that Differ in $m$ Operators} \label{sec:vary_m}
\begin{figure*}[htb]
    \centering
    \hspace*{-0.3in}\includegraphics[width=0.95\textwidth]{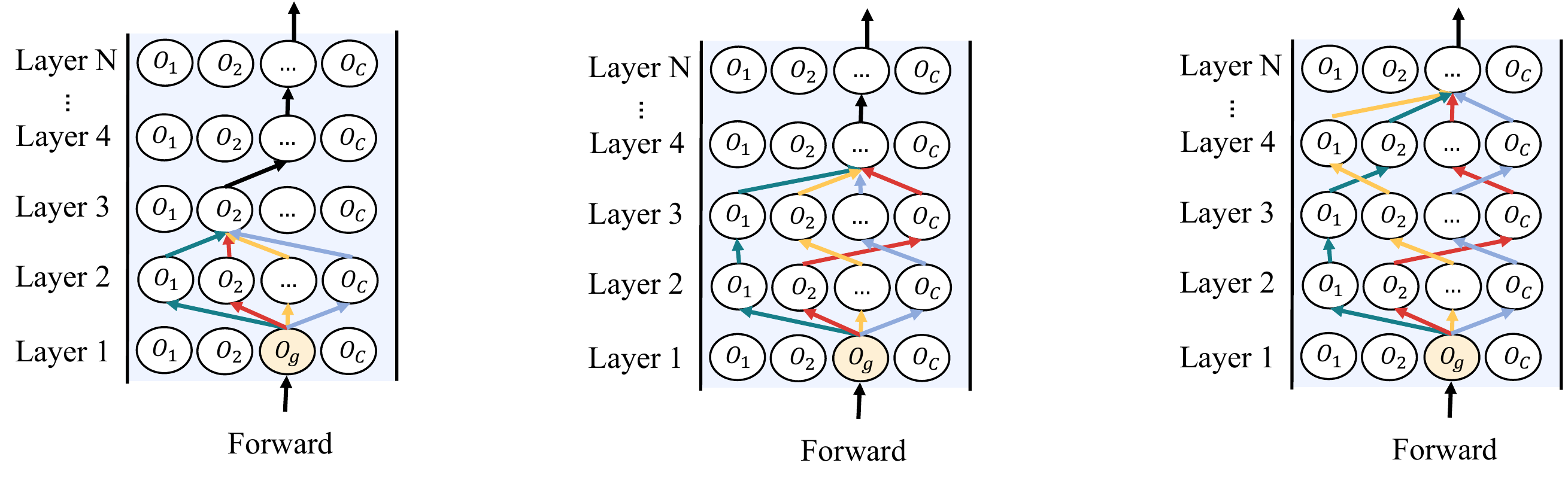}\\\vspace{-1mm}
  \begin{tabular}{ccc}
  \hspace*{0.2in}  (a) $m=1$  &  \hspace*{1.1in} (b) $m=2$ & \hspace*{1.2in}(c) $m=3$ \\
  \end{tabular}
\caption{The illustration of the forward process regarding the operator $o_g$ in layer 1 that is shared by child models that differ from the second layer to the $m+1$ layers.}
\label{fig:vary_m_demonstrate}
\end{figure*}
To extend the analyses to a more general case, we study the child models that differ in $m$ operators in Sec.~\ref{sec:pre_analyses_detail}. In this section, we depict how child models differ in $m$ operators to give readers a better understanding. Figure~\ref{fig:vary_m_demonstrate} (c) shows the case that we randomly choose $C$ child models that differ from the second layer to the fourth layer ($m=2$ different operators). Then the $C$ child models can apply the forward and backward computation individually and obtain their gradients on $o_g$. In this way, we can study the gradient interference on $o_g$ when child models differ in $m$ operators.

\begin{figure}[htb]
    \centering
    \includegraphics[width=0.3\textwidth]{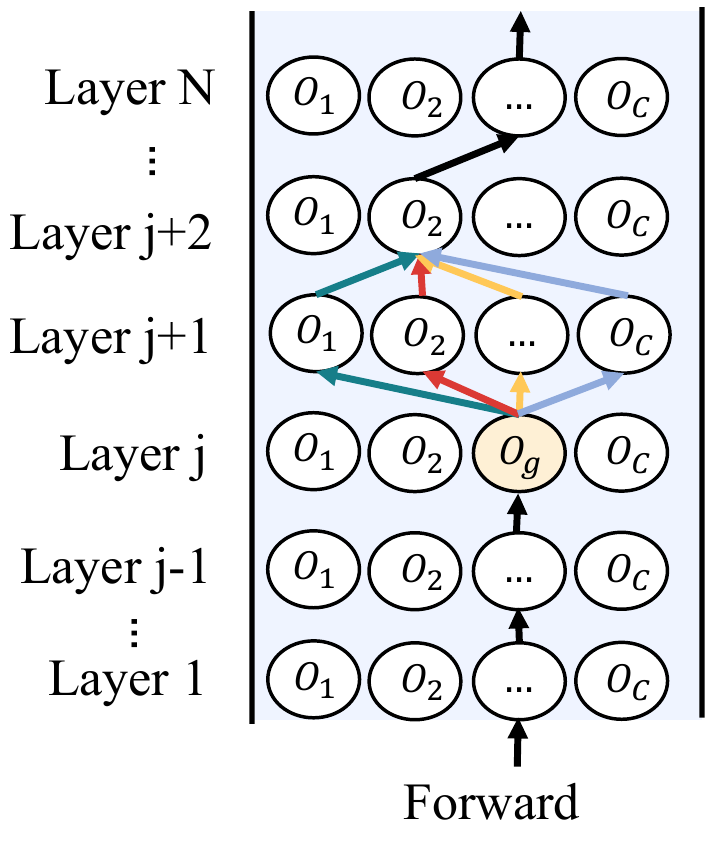}
    \vspace{-10pt}
\caption{The illustration of the forward and backward process regarding the operator $o_g$ in layer $j$ that is shared by child models that differ in layer $j+1$.}
\label{fig:o_g_differ_layer}
\end{figure}
\subsection{Study of Interference when $o_g$ Locates in Different Layers} \label{sec:o_g}



In the above analyses, we select $o_g$ in the first layer as an example to study interference. In this section,  we further investigate interference when $o_g$ locates in different layers. As shown in Figure~\ref{fig:o_g_differ_layer}, we choose $C$ child models which differ in layer $j+1$. Then we compare their gradients on $o_g$ at the $j$-th layer. The gradient cosine similarities with different $j$ are presented in Figure~\ref{fig:more_o_g}. We can find that our previous observations still hold no matter which layer $o_g$ locates in. Specifically, 1) by aligning the inputs and outputs of the shared operators to be similar to the average inputs and outputs, the cosine similarity is improved and the gradient interference can be reduced; 2) the interference on a shared operator between two child models is positively correlated with the number of different operators between them (the larger $m$, the lower cosine similarity).

\section{MAGIC-AT}
It is straightforward to combine MAGIC-A and MAGIC-T to collaboratively mitigate interference since these two independent methods proposed from two different perspectives are independent. Here we present the complete procedure for MAGIC-AT as follows:
\begin{itemize}[leftmargin=*]
    \item Obtain a batch of data, an anchor child model $\bm{\alpha}^l$ and a sampled child model $\bm{\alpha_{t-1}}$ at step $t-1$.
    \item Sample a child model $\alpha_t$ by randomly substituting one operator ($k=1$) in  $\bm{\alpha_{t-1}}$ with another operator,
    \item Calculate the loss according to Eq.~2 and update the weights of $\bm{\alpha_t}$,
    \item Replace $\bm{\alpha}^l$ with $\bm{\alpha_t}$ if $\text{Val}(\bm{\alpha}_t)>\text{Val}(\bm{\alpha}^l)$,
\end{itemize}
where $\text{Val}(\cdot)$ is the accuracy obtained from the validation set. 

\begin{figure}[htb]
\centering
    \includegraphics[width=0.25\textwidth]{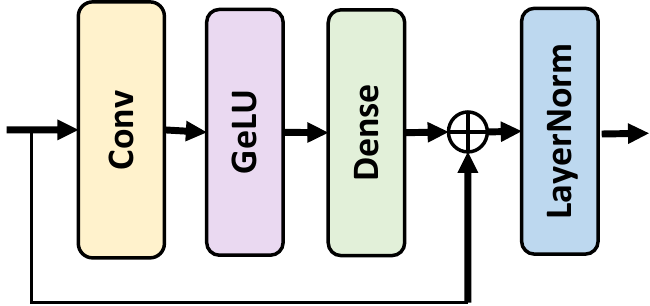}
\caption{Architecture of convolution operator.}
\vspace{-14pt}
\label{fig:conv_op}
\end{figure}

\begin{figure*}[tbh]
    \centering
    \hspace*{0.1in}\includegraphics[width=0.95\textwidth]{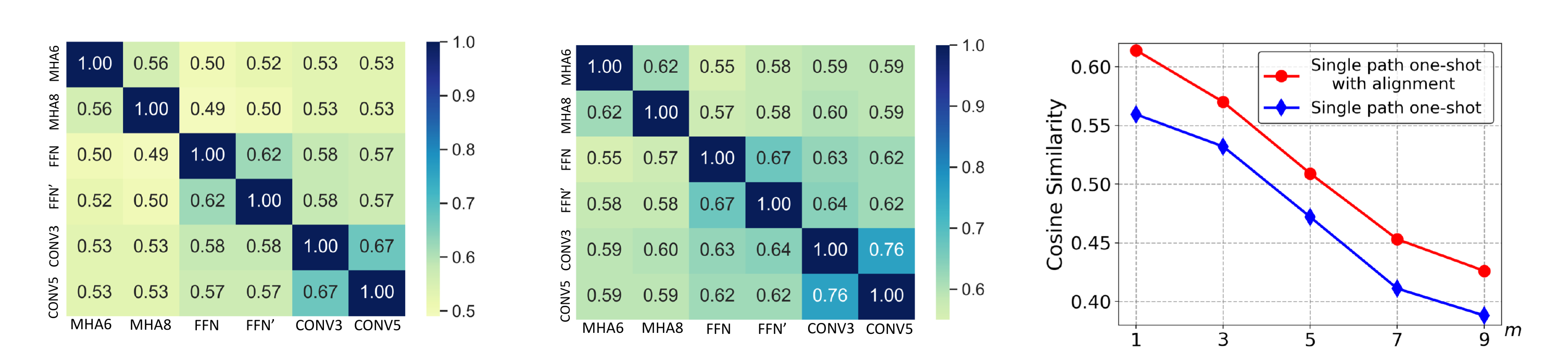}\\
  \small(a) Gradient cosine similarity comparison when $o_g$ locates in layer $j=3$\\
  \hspace*{0.1in}\includegraphics[width=0.95\textwidth]{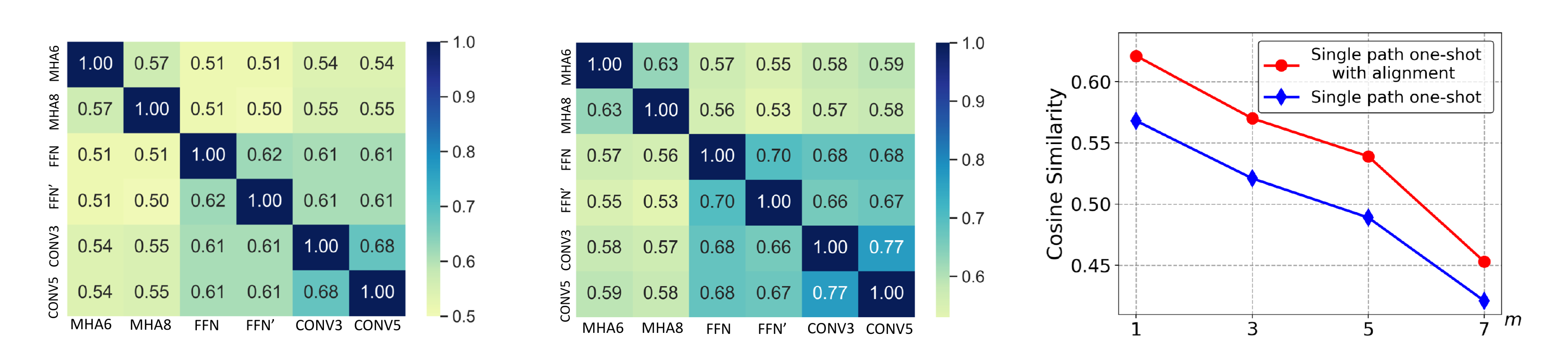}\\
  \small(b) Gradient cosine similarity comparison when $o_g$ locates in layer $j=5$\\
  \hspace*{0.1in}\includegraphics[width=0.95\textwidth]{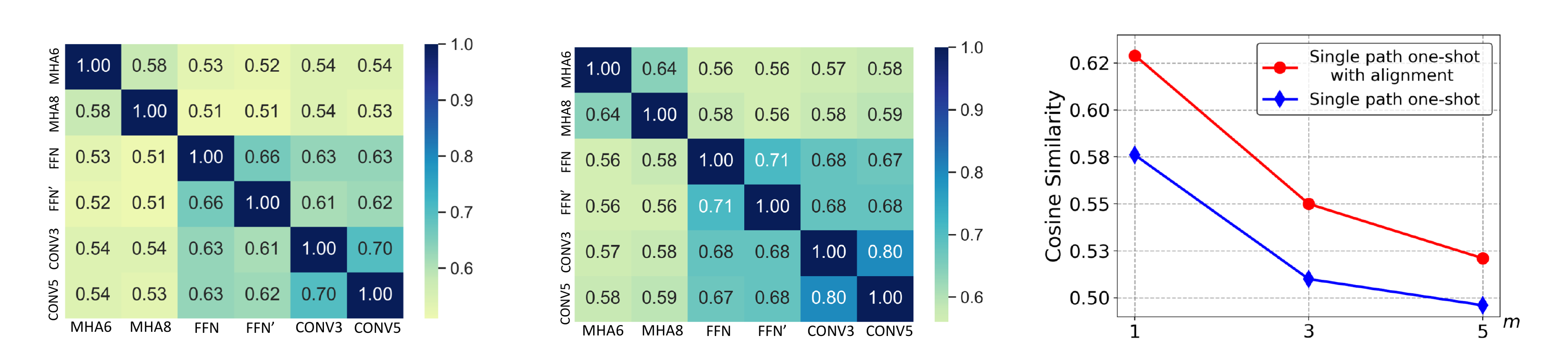}\\
  \small(c) Gradient cosine similarity comparison when $o_g$ locates in layer $j=7$\\
  \hspace*{0.1in}\includegraphics[width=0.95\textwidth]{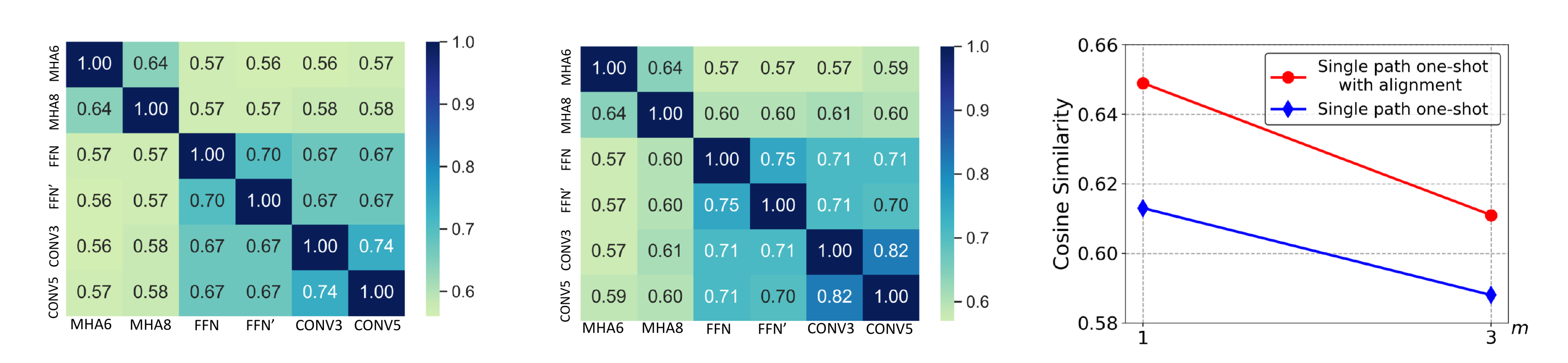}\\
  \small(d) Gradient cosine similarity comparison when $o_g$ locates in layer $j=9$\\
  \hspace*{0.1in}\includegraphics[width=0.95\textwidth]{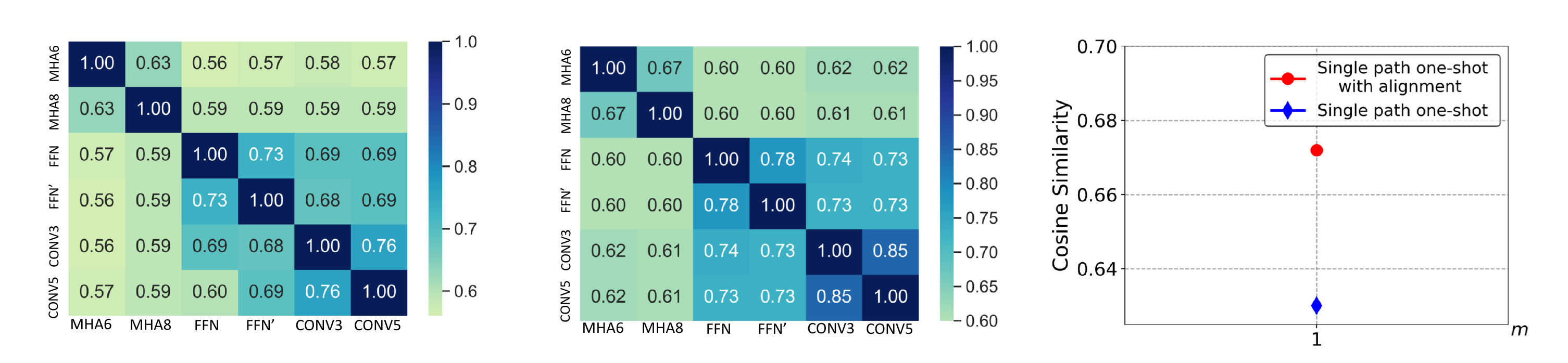}\\
  \small(e) Gradient cosine similarity comparison when $o_g$ locates in layer $j=11$\\
\caption{Gradient cosine similarity. By varying $j$, we repeat the experiments and present results as described in Sec.~3.2. }
\label{fig:more_o_g}
\end{figure*}

\section{Experiment Configurations}\label{appendix:arches_and_configs}
\paragraph{Candidate operators for BERT tasks.}
To search an effective architecture for practical usage, we adopt candidate operators $\mathcal{O}=$\{MHA12, FFN,  CONV3, CONV5\}, where FFN and MHA12 (head dimension 64) follow the designs in BERT$_{\rm base}$~\cite{devlin2019bert}, CONV with kernel size 3 and 5 are used since their effectiveness and efficiency in natural language tasks have been demonstrated by previous work~\cite{xu2021taskagnostic}. We adopt the architecture of CONV as shown in Figure~\ref{fig:conv_op} since we observe its superior performance and efficiency than that of the separable convolution. The hidden size of operators is 768 following BERT$_{\rm base}$~\cite{devlin2019bert}. BERT$_{\rm base}$ has 24 sub-layers (each Transformer layer has a MHA12 and FFN). Since CONV operators have fewer parameters than FFN, the searched architectures usually have fewer parameters than BERT$_{\rm base}$ with the same layers in the super-net. We adopt a chain-styled super-net with two more layers to enable that searched architecture can have similar parameters and FLOPs as BERT$_{\rm base}$. The space contains about 4.5$\times \text{10}^{\text{15}}$ child models in total.

\paragraph{Super-net Training}
We use Adam~\citep{KingmaB14} with a learning rate of 1e-4, $\beta_1=0.9$ and $\beta_2=0.999$. The peak learning rate is 5e-4 with a warmup step of 10,000 followed by linear annealing. The dropout rate is 0.1 and the weight decay is 0.01. We set the max length of sentences as 128 tokens. The super-net is trained with the batch size of 1024 sentences for 250,000 steps (same computational resources as BERT$_{\rm base}$). We adopt commonly-used SPOS~\cite{guo2020single} as the baseline method, which randomly samples a child model for training at each step. For our proposed method MAGIC-T, unless otherwise mentioned, it gradually changes $k=1$ operator to reduce the interference. For MAGIC-A, we train the super-net with masked language modeling loss in the first three epochs for a warm-start and then add the alignment loss with $\lambda=0.5$ in every four layers to avoid over-regularization. Our experiments are implemented with fairseq codebase~\cite{ott2019fairseq}.

\paragraph{GLUE Benchmark}
We evaluate performance by fine-tuning the pre-trained model on GLUE benchmark~\cite{WangSMHLB19}, which includes three inference tasks (MNLI~\citep{williams2018mnli}, QNLI~\citep{rajpurkar2016squad} and RTE~\citep{dagan2006rte}), three similarity and paraphrase tasks (MRPC~\citep{dolan2005mrpc} STS-B~\citep{cer2017stsb}, QQP~\citep{chen2018quora}) and two single-sentence tasks (CoLA~\citep{WarstadtSB19}, and SST-2~\citep{socher2013sst}). We follow hyper-parameters of RoBERTa  for fine-tuning, where STS-B, MRPC and RTE are started from the model fine-tuned on MNLI~\cite{liu2019roberta,ClarkLLM20,Song0QLL20}. 

\begin{figure*}[htbp]
\vspace{-1mm}
\centering
  \includegraphics[width=0.14\textwidth, angle=90]{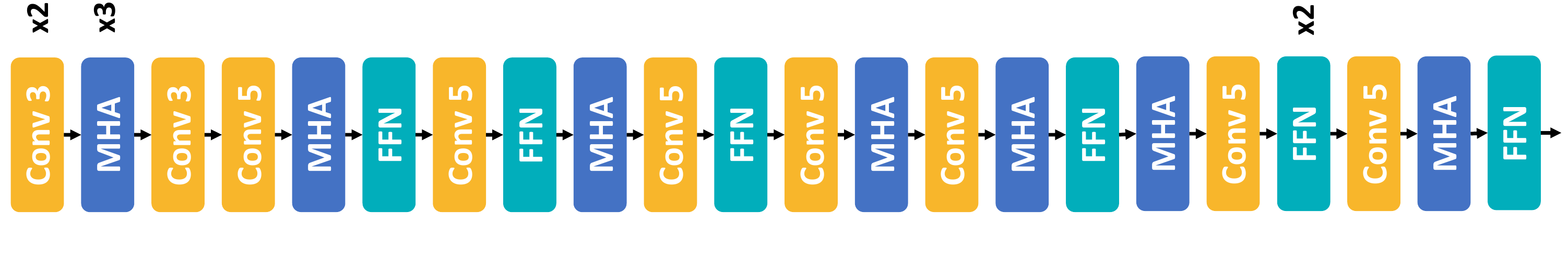}  
  \vspace{-2mm}
\caption{Architecture found by MAGIC-AT for the BERT pre-training task.}
\label{fig:arch_110M_pre}
\vspace{-3mm}
\end{figure*}

\paragraph{Configurations of the BERT pre-training task.}
To show the generality of the architecture found by our proposed MAGIC, we train the architecture with two different training objectives: 1) masked language modeling (MLM)~\cite{devlin2019bert} proposed in BERT~\cite{devlin2019bert} and 2) replace token detection (RTE) proposed in ELECTRA~\cite{ClarkLLM20}. For MLM training objective, RoBERTa~\cite{liu2019roberta} is a competitive baseline. For a fair comparison with it, we train our pre-training model with a large byte-level BPE vocabulary containing 50K subword units following RoBERTa~\cite{liu2019roberta}. However, a large vocabulary increases the size of parameters in the embedding layer. Thus, we factorize the embedding matrix into a multiplication of a small embedding matrix with hidden size 512 and another 512$\times$768 transformation matrix following~\cite{ALBERT2020}. For RTE training objective, we used the character-level BPE vocabulary of size 30K following ELECTRA~\cite{ClarkLLM20}. The searched architecture is shown in Figure~\ref{fig:arch_110M_pre}.

\begin{figure*}[htb]
\vspace{-2mm}
\centering
  \includegraphics[width=0.16\textwidth, angle=90]{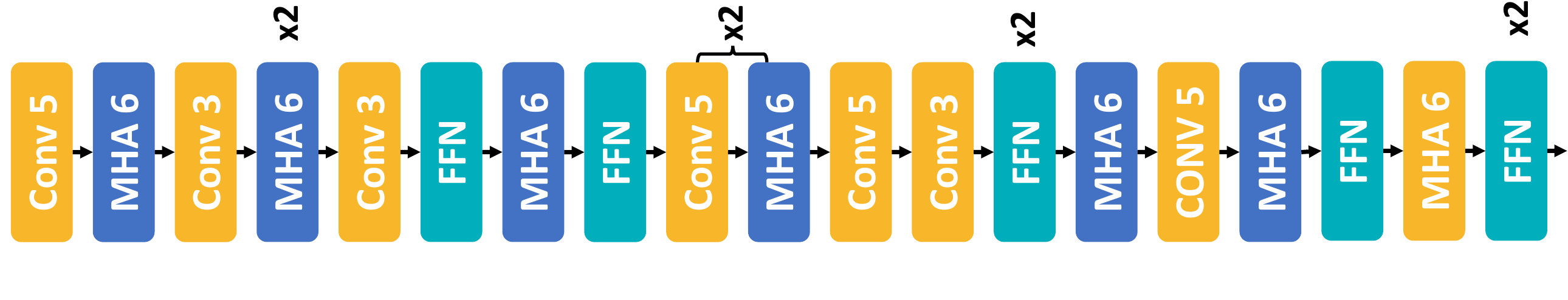}  
\caption{Architecture found by MAGIC-AT for BERT model compression.}
\label{fig:arch_110M_distl}
\end{figure*}

\paragraph{Configurations of the BERT compression task.}
Previous works usually compress BERT into a model size of 66M or 60M by taking advantage of novel architectures~\cite{xu2021taskagnostic} or sophisticated distillation techniques~\citep{sanh2019distilbert,wang2020minilm,turc2019well,hou2020dynabert}. Following NAS-BERT~\cite{xu2021taskagnostic}, we search a novel architecture of about 60M for comparison. To this end, we reduce the hidden size of candidate operators from 768 to 512 and re-run MAGIC-AT to search for architectures. The searched architecture is shown in Figure~\ref{fig:arch_110M_distl}. We adopt the large byte-level BPE vocabulary containing 50K subword units~\cite{liu2019roberta} and re-train the searched architecture exactly following NAS-BERT for a fair comparison.

\begin{figure*}[htb]
\vspace{-2mm}
\centering
  \includegraphics[width=0.9\textwidth]{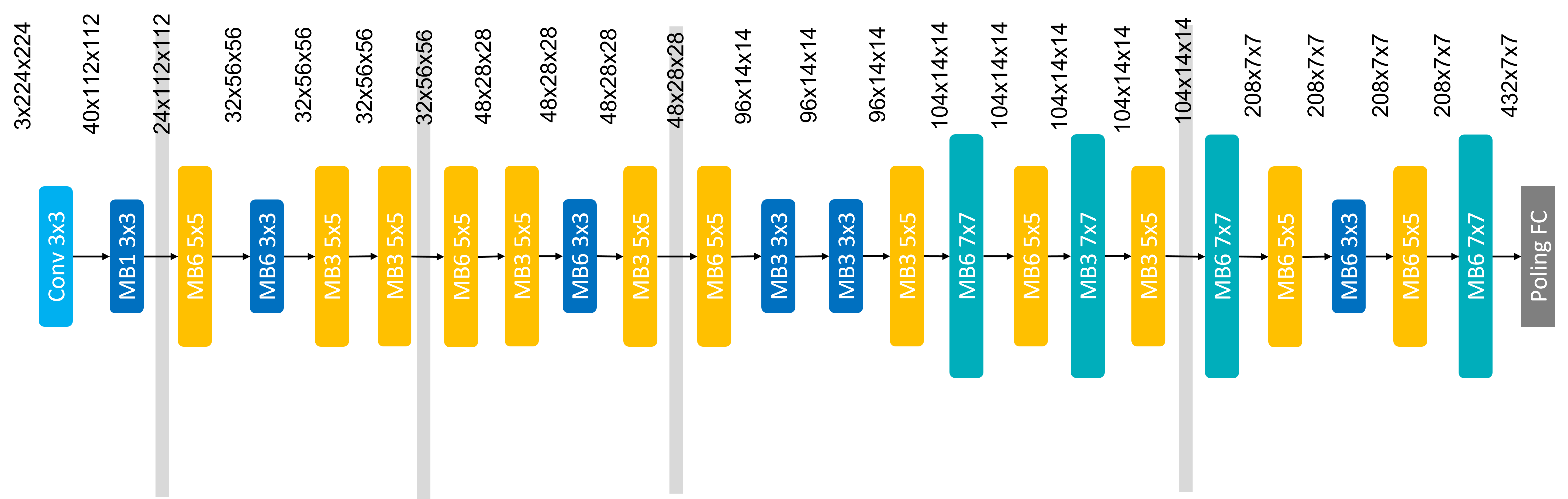}  
\caption{Architecture found by MAGIC-AT for ImageNet.}
\label{fig:arch_imagenet}
\end{figure*}

\paragraph{Configurations of the ImageNet task.} We adopt candidate operators $\mathcal{O}=$\{MB3 3$\times$3, MB3 6$\times$6, MB5 3$\times$3, MB5 6$\times$6, MB7 3$\times$3, MB7 6$\times$6, Zero-out\} following ProxylessNAS~\cite{cai2018proxylessnas}, where MB7 6$\times$6 refers to mobile inverted bottleneck convolution with kernel size 6 and expansion rate 7. We search the operation of each individual layer via MAGIC-AT. The discovered architecture by MAGIC-AT is shown in Figure~\ref{fig:arch_imagenet}.

\end{document}